\documentclass{article} 
\usepackage{iclr2021_conference,times}

\usepackage{amsmath,amsfonts,bm}

\def\eqref#1{equation~\ref{#1}}

\def\1{\bm{1}}

\DeclareMathAlphabet{\mathsfit}{\encodingdefault}{\sfdefault}{m}{sl}
\SetMathAlphabet{\mathsfit}{bold}{\encodingdefault}{\sfdefault}{bx}{n}

\usepackage{hyperref}
\usepackage{url}

\usepackage{blindtext}
\usepackage{kotex}
\usepackage{xcolor}
\usepackage{graphicx}
\usepackage{multirow}
\usepackage{caption}
\usepackage{subcaption}
\usepackage{float}

\definecolor{cadmiumgreen}{rgb}{0.0, 0.42, 0.24}
\definecolor{cadmiumred}{rgb}{0.89, 0.0, 0.13}

\title{What is Wrong with\\One-Class Anomaly Detection?}

\author{Junekyu Park, Jeong-Hyeon Moon, Namhyuk Ahn \& Kyung-Ah Sohn \\
Department of Artificial Intelligence, Ajou University\\
\texttt{idbluefish@gmail.com, \{mjh319,aa0dfg,kasohn\}@ajou.ac.kr} \\
}

\iclrfinalcopy 
\begin{document}

\maketitle

\begin{abstract}

From a safety perspective, a machine learning method embedded in real-world applications is required to distinguish irregular situations.
For this reason, there has been a growing interest in the anomaly detection (AD) task.
Since we cannot observe abnormal samples for most of the cases, recent AD methods attempt to formulate it as a task of classifying whether the sample is normal or not.
However, they potentially fail when the given normal samples are inherited from diverse semantic labels.
To tackle this problem, we introduce a latent class-condition-based AD scenario.
In addition, we propose a confidence-based self-labeling AD framework tailored to our proposed scenario.
Since our method leverages the hidden class information, it successfully avoids generating the undesirable loose decision region that one-class methods suffer.
Our proposed framework outperforms the recent one-class AD methods in the latent multi-class scenarios.

\end{abstract}

\section{Introduction}
With the rapid increase in the performance of deep learning-based methods, the demands for applying this technology are emerging in recent.
However, simply adopting it may not be ideal due to the mismatched label information between the training and test set.
A self-driving system, for example, should make the best decision on the abnormal scene or status even though it never observed this condition before.
Under this scenario, it is required to detect whether the given data is unseen (abnormal) or not, to build a reliable and secure machine learning system.

The anomaly detection (AD) task focuses on to identify suspicions or abnormal events.
We can categorize this task into supervised~\citep{liang2017enhancing} or unsupervised~\citep{chalapathy2017robust,oza2018one} by the training strategy.
The former train the model with both normal and abnormal samples, while the latter use normal data only.
Since collecting the abnormal cases is time-consuming or impossible in the real-world, unsupervised learning gets more attention despite the superior performance of the supervised scheme.
In an unsupervised approach, most of the methods formulate the task as an one-class classification problem (\textit{i.e.} classify as normal or not).
While such simplicity works on the well-refined scenarios, they potentially fail when the given normal dataset is composed of samples from diverse classes (Figure~\ref{fig:overview_prev}).
Since conventional methods ignore the latent class information, they tend to draw a single binary decision boundary that loosely covering a wide range.
This could be problematic when the normal dataset contains multiple latent class information.
Can we leverage this semantic knowledge to guide the AD methods to judge the anomalies more accurately? We believe a different approach is needed.

Base on this assumption, we introduce a latent class-condition-based AD scenario and its benchmark datasets.
This simulates the circumstance where both normal and abnormal data have multi-class samples (Figure~\ref{fig:overview_ours}).
We would like to emphasize that our proposed scenario is realistic; in the real-world, various (object) classes could be normal, and the abnormal cases could consist of diverse classes as well (\textit{e.g.} street signs of different countries).
Because of the nature of our scenario, the optimal solution needs to generate semantic-aware \textit{tight} decision boundaries (as in Figure~\ref{fig:overview_ours}).
However, neither supervised nor unsupervised approaches can handle the scenario properly.
For example, supervised models cannot be applied since no label is given. Conventional unsupervised AD methods create a single and loose decision region (since they ignore latent class information).
At this point, one natural question can arise: How can we use latent semantic information?
Since the latent label is accessible to the oracle (or human) only, we may detour this limitation, for example, by utilizing the pseudo-latent class label.

\begin{figure*}[t]
\centering
\includegraphics[width=\linewidth]{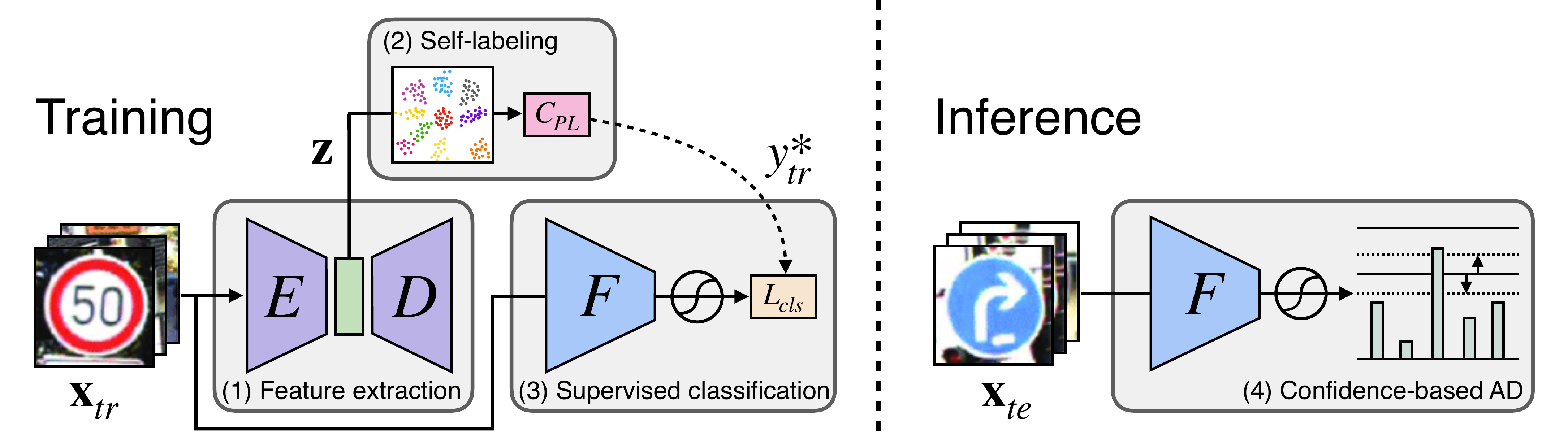}
\caption{
Overview of our proposed confidence-based self-labeling anomaly detection framework.
\textbf{(1)} Feature extraction by training the convolutional autoencoder to extract the latent feature $\mathbf{z}$.
\textbf{(2)} Self-labeling via clustering to assign latent class ${y}^{*}$ to sample $\mathbf{x}$.
\textbf{(3)} Train a classifier $F$ using $(X_{tr}, Y^{*}_{tr})$ pairs in supervised learning manner. $L_{cls}$ denotes a classification loss.
\textbf{(4)} We perform anomaly detection by the confidence score of classifying normal sample $\mathbf{x}$ to $y^*$.
}
\label{fig:model}
\end{figure*}

To tackle this, we propose a Confidence-based self-Labeling Anomaly Detection (CLAD) framework to bridge the gap between the supervised and unsupervised approaches.
We model the latent class information using self-labeling so that supervised learning can be adapted (Figure~\ref{fig:model}).
To do this, we first train the feature extraction network following~\cite{xie2016unsupervised}.
Then, we cluster the training samples using the extracted latent feature and allocate pseudo labels via self-labeling~\citep{lee2013pseudo}.
Now the classification network can learn to predict pseudo labels by standard supervised learning.
At the inference, we decide whether the test sample is abnormal or not by the confidence-based AD method~\citep{liang2017enhancing}.
Since our method leverages the hidden class information, it successfully avoids the generation of undesirable loose decision regions typically suffered by one-class methods.
Several experiments on latent multi-class scenarios demonstrate that the proposed method substantially outperforms recent one-class AD methods.\footnote{Our code is available at \url{https://github.com/JuneKyu/CLAD}}

\section{Latent Class-condition Anomaly Detection Scenario}
\begin{figure*}[t]
\centering
\setlength{\tabcolsep}{0.5em}
\begin{tabular}{cc}
   \subfloat[One-class AD\label{fig:overview_prev}]{\includegraphics[width=0.475\linewidth]{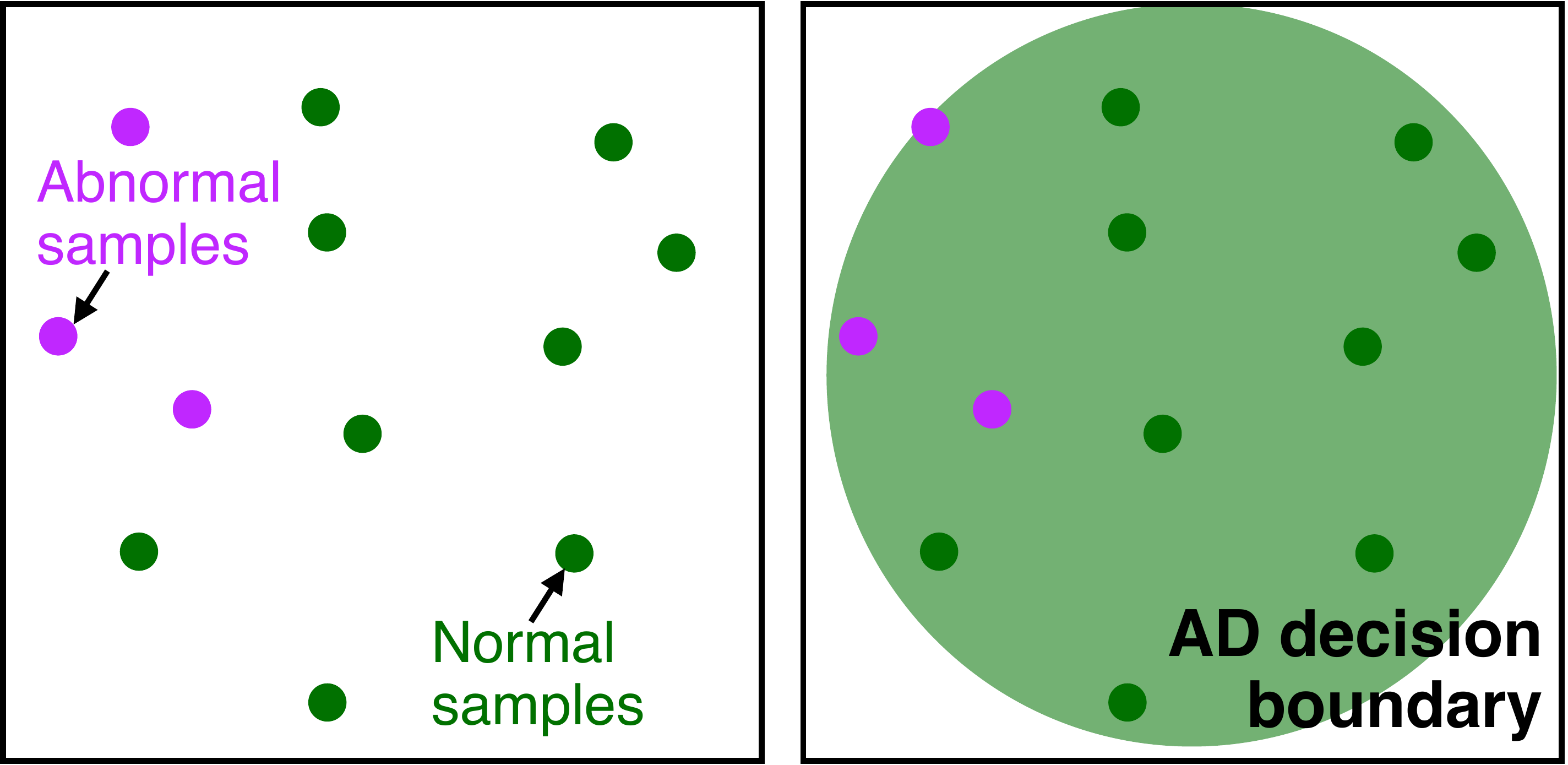}}  & 
   \subfloat[Our proposed AD\label{fig:overview_ours}]{\includegraphics[width=0.475\linewidth]{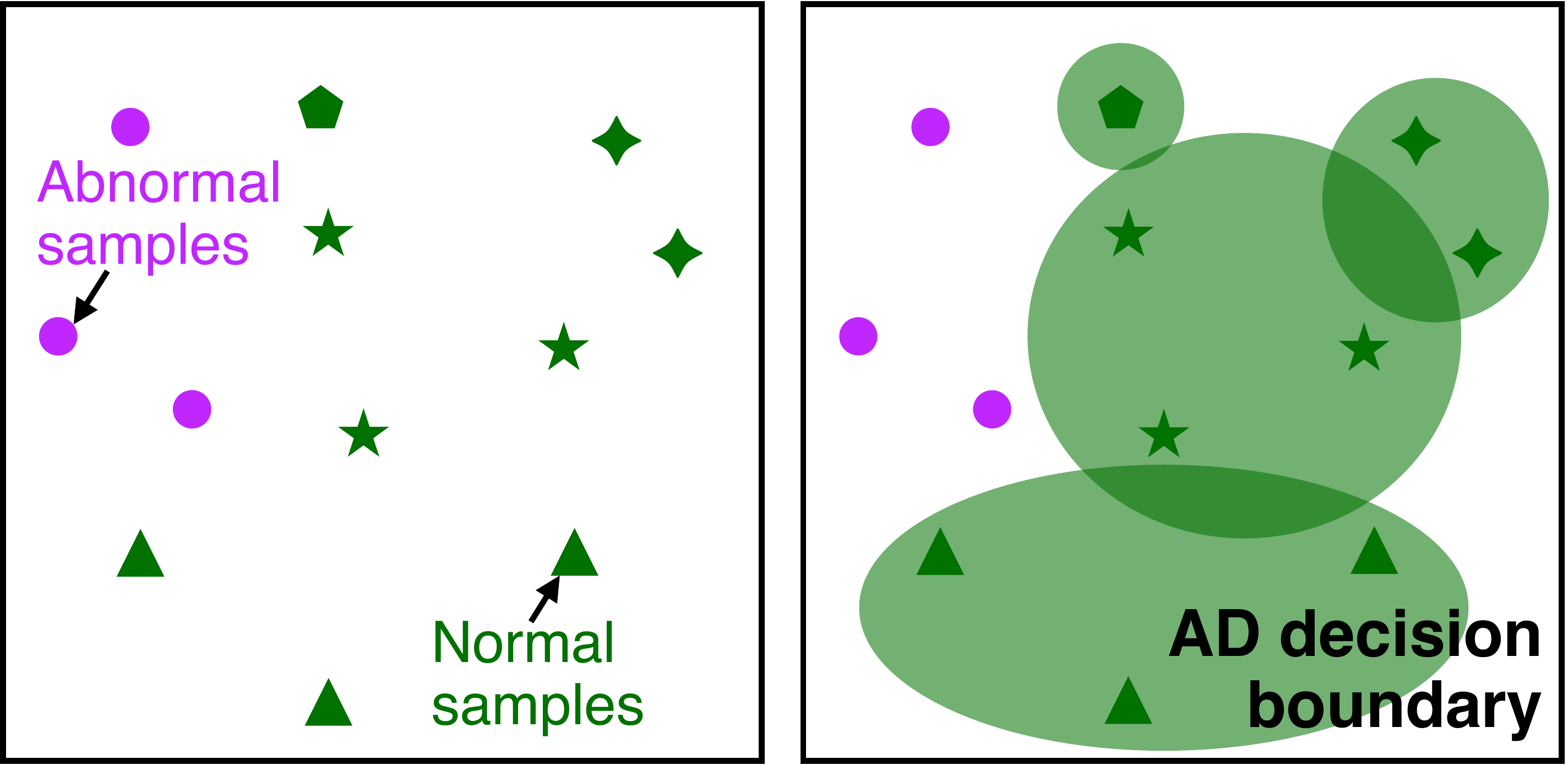}}
\end{tabular}
\caption{Comparison of two AD scenarios. \textbf{(a)} One-class AD scenario where the model creates a single decision boundary to cover all normal data. Some abnormal samples are misclassified due to the \textit{loose} boundary. \textbf{(b)} Our proposed AD scenario. An optimal solution can model latent class-condition information and draws \textit{tight} class-wise decision boundaries. For normal samples, each class is denoted as different shapes. Note that the latent class information are not observable to the models; only the oracle is able to access this.}
\label{fig:overview}
\end{figure*}

Traditional one-class AD methods learn a decision boundary based on the given normal samples in the training dataset (Figure~\ref{fig:overview_prev}).
Such one-class strategy is effective when a single class label is treated as normal alone (and the rest of them are abnormal).
However, this assumption may not hold in a real environment since the normal category can consist of heterogeneous semantic labels.
With this circumstance, conventional one-class AD generates a loose decision boundary to cover all samples drawn from diverse classes and thus vulnerable to a false negative.

To reduce the gap between the real-world and the one-class AD scenarios, we simulate the scenario environment where the latent sub-classes exist implicitly (Figure~\ref{fig:overview_ours}).
With this environment, it is crucial to learn a decision boundary by seeing not only the normality of the data samples but also its semantics.
Note that such class information is not observable, thus the AD framework may require learning the semantic representation in an unsupervised or self-supervised manner.

\section{Confidence-based Self-labeling Anomaly Detection}

Before going into the details, let us define the problem statement.
We have a training set $X_{tr} = \{\mathbf{x}_i\}^{N}_{i=1}$ and a test set $(X_{te},Y_{te})$. $X_{te}$ contains both 
normal/abnormal samples with corresponding label $Y_{te}$; $0$ if normal and $1$ otherwise.
The core idea of our framework is to consider the concealed semantic information.
To do that, we generate the pseudo-label $y^*  \in  \mathcal{Y}^* = \{1,...,L\}$, where the cardinality of $\mathcal{Y}^{*}$ is assumed pseudo-class counts.
Then, our AD framework inferences a true $y \in Y_{te}$ based on the classifier $F$ that is trained with generated pairs $(X_{tr}, Y^*_{tr})$, where $Y^*_{tr} = \{\mathbf{y}^*_i\}^{N}_{i=1}$.

\noindent\textbf{(1) Feature extraction.}
First, we train the autoencoder ($E$ and $D$) with reconstruction loss on training dataset $X_{tr}$ and calculate the latent feature $\mathbf{z}$ using the trained encoder $E$.

\noindent\textbf{(2) Self-labeling via clustering.} 
We further fine-tune the encoder $E$ to increase the possibility of assigning a sample $\mathbf{x}$ to the given cluster $C_i$ by the Kullback-Leibler divergence between the latent feature $\mathbf{z}$ and soft-alignment of $C_i$~\citep{xie2016unsupervised}.
We simultaneously update the encoder and cluster assignment and this makes the clustering algorithm to be robust.
Then, we assign a label $y^*$ by referring to the allocated cluster $C_i$.
Now the $X_{tr}$ have associated pseudo-labels ${Y}_{tr}^{*}$.

\noindent\textbf{(3) Supervised classification.}
With (pseudo) latent class labels $Y^*_{tr}$ and the normal sample $X_{tr}$, a classifier $F$ is trained with a standard supervised learning scheme.
By referring to the class information that is intrinsic in the latent labels, our framework can generate the semantic-aware \textit{tight} decision boundaries that envelop each relevant class sample only.

\noindent\textbf{(4) Confidence-based AD.}
Because of the tight decision boundary by the pairs $(X_{tr}, Y^*_{tr})$, we argue that the AD can be viewed as the out-of-distribution (OOD) task.
In detail, the OOD sample is defined as the data where the class label is not included in the training dataset. 
We assume the label set $\mathcal{Y}^*$ is a multi-classes label set from $Y^*_{tr}$.
Then, we treat $(\mathbf{x}, y^*)$ as an OOD sample when the $y^* \not\in \mathcal{Y}^*_{tr}$.
Because the confidence-based algorithms predict the samples as OOD when the classifier outputs a small probability for all classes, we can safely convert anomaly detection into an OOD task.
(\textit{i.e.}, it is an abnormal sample when $y^*\not\in\mathcal{Y}^*_{tr}$).

Under these assumptions, we measure anomaly detection scores by adapting the scoring scheme in ODIN~\citep{liang2017enhancing}.
We define the score $s(\mathbf{x};T,\delta) =max_i p(\tilde{\mathbf{x}};T)_{(i)}$, where $p(\tilde{\mathbf{x}};T)_{(i)}$ is output of the classifier $F$ in each class $i$.
$T$ is the parameter for the temperature scaling and $\tilde{\mathbf{x}}$ is the input term $\tilde{\mathbf{x}} = \mathbf{x} - \epsilon(\mathbf{x})$ 
perturbed by the reverse FGSM method~\citep{goodfellow2014explaining} that makes the OOD samples more separable.
If the score $s(x;T,\delta)$ is greater than a given threshold $\delta$, then we predict the $\hat{y}$ as normal.

\section{Experiment}

\noindent\textbf{Baselines.}
We compare with one-class AD methods: OCSVM~\citep{scholkopf2001estimating}, OCNN~\citep{chalapathy2017robust}, OCCNN~\citep{oza2018one}, SVDD~\citep{tax2004support}, and DeepSVDD~\citep{ruff2018deep}.
We follow the implementation setups based on the official codes.

\noindent\textbf{Datasets.}
We use following datasets: MNIST~\citep{lecun-mnisthandwrittendigit-2010}, CIFAR-10~\citep{krizhevsky2009learning}, GTSRB~\citep{Stallkamp2012}, and Tiny-ImageNet~\citep{ILSVRC15}.

We devise the super-categories by merging the semantic labels to simulate our AD scenario.
For example, MNIST is as \{\texttt{Curly, Straight, Mix}\} and GTSRB based on the semantic meanings of traffic signs.
Note that both datasets share a similar domain prior which represents the text or symbols.
For the natural scene datasets such as CIFAR-10 and Tiny-ImageNet, we set as \{\texttt{Thing, Living}\} and \{\texttt{Animal, Insect, Instrument, Structure, Vehicle}\}, respectively.
When we evaluate the models, we pick one super-category for training and the rest of the subsets as a test dataset.
Please see Appendix \ref{sec:scenario_stup} for the detailed settings of the scenario.

\noindent\textbf{Results.}
Table~\ref{mnist_performance} shows the AUROC of the MNIST and GTSRB datasets.
Our framework surpasses all the one-class AD methods in MNIST.
Among them, it is notable that CLAD outperforms others on \texttt{Mixed} in a huge margin.
This scenario is very challenging since it carries complex information due to the mixed shape of digits such as `2' or `6'.
For the GTSRB dataset, our CLAD reaches the best performance for \texttt{SPDL, DIRC, REGN} and second-best for the rest.
The only method comparable to ours is DeepSVDD. However, it suffers the inconsistent performance (worst score in \texttt{SPEC}) while our framework shows stable and high performances for all the cases.

To demonstrate the superior performance of CLAD in the complex image domain, we evaluate it on the CIFAR-10 and Tiny-ImageNet datasets (Table~\ref{cifar10_performance}).
Our framework achieves the best performance for all the scenarios in CIFAR-10 and four of five cases in Tiny-ImageNet.
Similar to GTSRB, the scores of DeepSVDD are inconsistent; we claim that DeepSVDD is sensitive to the latent class labels.
It maps all the (normal) samples into a single-modal hypersphere (Figure~\ref{fig:overview_prev}), making it vulnerable to the anomalies that close to the normals in terms of the class information.

\begin{table}[t]
\caption{Performance comparison on the newly proposed AD scenario with MNIST and GTSRB.}\label{mnist_performance}
\begin{center}
\begin{tabular}{l|ccc|cccccc}
\hline
\multirow{2}{*}{Method}  & \multicolumn{3}{c|}{\bf MNIST}  & \multicolumn{6}{c}{\bf GTSRB} \\
          \cline{2-10}
       & CUR & STR & MIX & SPDL & INST & WARN & DIRC & SPEC & REGN \\
          \hline\hline
OCSVM  & 69.9 & \underline{87.4} & 54.8 & \underline{66.3} & 60.9 & 52.7 & 60.7 & 50.5 & 77.2 \\
OCNN   & 81.4 & 76.2 & 60.2 & 65.7 & 65.7 & 51.8 & \underline{62.0} & 52.7 & \underline{78.3} \\
OCCNN  & 77.5 & 78.3 & 51.5 & 59.3 & 57.5 & 60.4 & 56.4 & 59.1 & 64.3 \\
SVDD   & 58.8 & 72.4 & 52.6 & 58.2 & 55.1 & 51.1 & 50.5 & 56.9 & 67.4 \\
DeepSVDD  & \underline{81.7} & 82.2 & \underline{69.7} & 57.7 & \textbf{69.7} & \textbf{73.4} & 59.3 & \textbf{70.2} & 77.8 \\
\hline
CLAD (ours) & \bf 94.0 & \bf 96.1 & \bf 92.6 & \bf 66.7 & \underline{66.5} & \underline{64.9} & \bf 67.4 & \underline{62.2} & \bf 79.1
\\
\hline
\end{tabular}
\end{center}
\end{table}

\begin{table}[t]
\caption{Performance comparison on our AD scenario with CIFAR-10 and Tiny-ImageNet.}\label{cifar10_performance}
\begin{center}
\begin{tabular}{l|cc|ccccc}
\hline
\multirow{2}{*}{Method} & \multicolumn{2}{c|}{\bf CIFAR-10} & \multicolumn{5}{c}{\bf Tiny-ImageNet} \\
          \cline{2-8}
         & THG & LIV & ANML & ISCT & ISTM & STRT & VHCL \\
          \hline\hline
OCSVM    & 51.9 & \underline{67.7} & \underline{63.5} & 60.9 & 50.2 & \underline{56.7} & 55.5 \\
OCNN     & 58.5 & 67.2 & 58.2 & 57.1 & 50.3 & 51.0 & 55.1 \\
OCCNN    & 59.8 & 62.5 & 59.7 & \underline{62.4} & 51.5 & 54.2 & \textbf{66.4} \\
SVDD     & 50.8 & 61.4 & 51.5 & 51.8 & 51.5 & 50.6 & 51.9 \\
DeepSVDD & \underline{65.0} & 52.7 & 59.0 & 53.5 & \underline{53.4} & 55.4 & 53.5 \\
\hline
CLAD (ours) & \bf 74.9 & \bf 72.8 & \bf 65.9 & \bf 66.2 & \bf 55.6 & \bf 62.0 & \underline{64.7} \\
\hline
\end{tabular}
\end{center}
\end{table}

\section{Discussion \& Conclusion}
We introduced a new aspect of the AD task and proposed a confidence-based AD framework.
We assume that (ab)normal categories can have (unobservable) multi-class samples in contrast to the one-class AD scenarios.
We believe that our scenario and method are practical in real world. 
One possible usage is for the industrial environment.
When the \textit{normal} sensor signals with various range can be altered by the surroundings, conventional one-class AD methods suffer spurious detection.

Since our framework trains a classifier using self-labeling, extracting the right representation of the data sample is crucial.
However, most of the feature extraction methods have difficulty with handling a dataset bias~\citep{bahng2020learning}.
As future work, we will focus on the disentanglement of the scene context with a key concept of the object discovery~\citep{burgess2019monet}.

\section*{Acknowledgements}
This research was supported by the National Research Foundation of Korea grant funded by the Korea government (MSIT) (No. NRF-2019R1A2C1006608), and also under the ITRC (Information Technology Research Center) support program (IITP-2020-2018-0-01431) supervised by the IITP (Institute for Information \& Communications Technology Planning \& Evaluation).

\bibliographystyle{iclr2021_conference}
\bibliography{}

\newpage
\appendix

\section{Related Work}
\textbf{One-class anomaly detection.}
Most of the anomaly detection (AD) methods treat a task as a binary classification: normal and abnormal.
Under this assumption, OCSVM~\citep{scholkopf2001estimating} learns a decision boundary with normal samples only.
OC-NN and OC-CNN~\citep{chalapathy2017robust, oza2018one} are the earlier attempts on using a deep learning-based approach with an unsupervised learning regime.
However, unlike OCSVM, these works have the potential risk to be a trivial solution due to the insufficient theoretical analysis.
That is, the learnable parameters may become a trivial solution when the given data samples are all normal.
On the other hand, SVDD~\citep{tax2004support} and DeepSVDD~\citep{ruff2018deep} successfully avoid such trivialness by the theoretical basis.
Although the aforementioned methods have shown promising results, they are limited to the one-class scenario which is not suitable for real-world applications.
In contrast, our proposed task assumes that the data points could be allocated into the various classes, which is more realistic.

\textbf{Self-labeling.}
The self-labeling is one of the most rapidly developing approaches in the self-supervised learning context~\citep{dosovitskiy2015discriminative, doersch2015unsupervised, noroozi2016unsupervised, noroozi2017representation, doersch2017multi, gidaris2018unsupervised}.
Base on the success of self-supervised learning, recent works attempt to create a self-label using traditional clustering methods. 
For example, DeepCluster~\citep{caron2018deep} first makes initial self-labels using a convolutional network and iteratively updates the network parameters with the re-assigning process by the clustering algorithms.
\cite{asano2019self} extended DeepCluster making it to learn visual representation and clustering simultaneously based on the information theory as such maximizing the information between the labels and the input data.
Although self-labeling has shown outstanding performance in the computer vision field, to the best of our knowledge, none of the studies exists to apply this strategy to the anomaly detection field.

\textbf{Out-of-distribution detection.}
Out-of-distribution (OOD) detection is the task to discriminate the samples whether they are from the training distribution or not.
Because the deep learning-based classification model tends to predict as a wrong class with high-confidence when the given test samples are the class not in the training set (\textit{i.e.} high-confidence problem).
To tackle this issue, many works have been studied on this problem.
ODIN~\citep{liang2017enhancing} addressed the high-confidence problem using temperature scaling and an input-preprocessing method from the FGSM~\citep{goodfellow2014explaining}.
Their method pursues better separating the out-of- and in-distribution samples. 
\cite{papadopoulos2019outlier} proposed an additional loss function from ODIN and applied other machine learning tasks such as natural language problems. 
The aforementioned methods rely on the multi-class labels when training, \textit{i.e.} supervised approach.
This limits the OOD detection methods difficult to apply to real-world anomaly detection problems.

\section{Model Analysis}
In this section, we analyze our proposed method.
First, we evaluate CLAD on the previous one-class AD scenario.
Second, we show how our method is robust to the hyper-parameter settings such as the number of the clusters or the hidden dimension size.

\textbf{One-class AD.} We can view this task as a simplification of our proposed scenario.
In one-class AD, the normal category has a single semantic label only, in contrast to ours which sets the normal condition to contain multiple class information.
We evaluate the methods on MNIST and CIFAR-10 datasets as shown in Table~\ref{mnist_one_class} and~\ref{cifar10_one_class}.
Our CLAD shows comparable results on MNIST to the other one-class-based AD methods and competes on par with DeepSVDD on the CIFAR-10 dataset.

We would like to note that CLAD is not designed for the one-class AD task.
Because of the feature clustering and label assignment, our method could create fragmented decision boundaries in this conventional scenario.
In contrast, DeepSVDD learns to generate a spherical decision boundary that tightly wraps the single-class normal samples.
However, on CIFAR-10 which may have various latent semantics within the single class (\textit{e.g.} different pose, intra-class diversity), DeepSVDD fails to detect abnormal in some cases (\textit{e.g.} bird, deer) while CLAD shows consistent scores.

\begin{table}[t]
\caption{One-class AD performance on MNIST dataset.}\label{mnist_one_class}
\begin{center}
\begin{tabular}{l|cccccccccc}
\hline
          \multirow{2}{*}{Method} & \multicolumn{10}{c}{\bf MNIST} \\
          \cline{2-11}
          & 0    & 1    & 2    & 3    & 4    & 5    & 6    & 7    & 8    & 9 \\
          \hline\hline
OCSVM     & \underline{98.3} & 99.5 & 82.0 & \underline{88.5} & 91.6 & 79.7 & 93.1 & 93.4 & 83.9 & 91.3 \\
OCNN      & 97.6 & \underline{99.5} & 87.3 & 86.5 & \underline{93.3} & 86.5 & \underline{97.1} & 93.6 & 88.5 & \underline{93.5} \\
OCCNN     & 91.8 & 98.7 & 74.9 & 78.2 & 88.3 & 72.7 & 75.6 & 85.6 & 69.1 & 78.5 \\
SVDD      & \bf98.6 & \underline{99.5} & 82.5 & 88.1 & \bf94.9 & 77.1 & 96.5 & \underline{93.7} & \underline{88.9} & 93.1 \\
DeepSVDD  & 98.0 & \bf99.7 & \bf91.7 & \bf91.9 & \bf94.9 & \underline{88.5} & \bf98.3 & \bf94.6 & \bf93.9 & \bf96.5 \\
\hline
CLAD (ours) & 96.3 & 97.9 & \underline{89.8} & 87.2 & 92.2 & \bf90.7 & 92.5 & 91.5 & 80.0 & 92.0 \\
\hline
\end{tabular}
\end{center}
\end{table}

\begin{table}[t]
\caption{One-class AD performance on CIFAR-10 dataset.}\label{cifar10_one_class}
\begin{center}
\begin{tabular}{l|cccccccccc}
\hline
          \multirow{2}{*}{Method} & \multicolumn{10}{c}{\bf CIFAR-10} \\
          \cline{2-11}
          & \footnotesize{airplane} & \footnotesize{automobile} & \footnotesize{bird} & \footnotesize{cat} & \footnotesize{deer} & \footnotesize{dog} & \footnotesize{frog} & \footnotesize{horse} & \footnotesize{ship} & \footnotesize{truck} \\
          \hline\hline
OCSVM     & \underline{65.1} & 59.0 & \bf65.2 & 50.1 & \bf75.1 & 51.3 & \underline{71.7} & 51.2 & 67.6 & 51.0 \\
OCNN      & 60.4 & 62.0 & \underline{63.7} & 53.6 & 67.4 & 56.1 & 63.3 & 60.1 & 64.7 & 60.3 \\
OCCNN     & 65.0 & \underline{65.6} & 62.8 & 51.2 & 72.7 & 50.9 & 64.7 & 52.2 & 66.5 & 66.9 \\
SVDD      & 61.6 & 63.8 & 50.0 & 55.9 & 66.0 & 62.4 & 74.7 & \underline{62.6} & \underline{74.9} & \bf75.9 \\
DeepSVDD  & 61.7 & \bf65.9 & 50.8 & \underline{59.1} & 60.9 & \bf65.7 & 67.7 & \bf67.3 & \bf75.9 & \underline{73.1} \\
\hline
CLAD (ours) & \bf73.0 & 64.4 & 58.3 & \bf60.1 & \underline{73.1} & \underline{63.9} & \bf76.0 & 60.1 & 70.7 & 69.8 \\
\hline
\end{tabular}
\end{center}
\end{table}

\begin{figure}[t]
\centering
\begin{subfigure}[b]{0.48\textwidth}
    \centering
    \includegraphics[width=\textwidth]{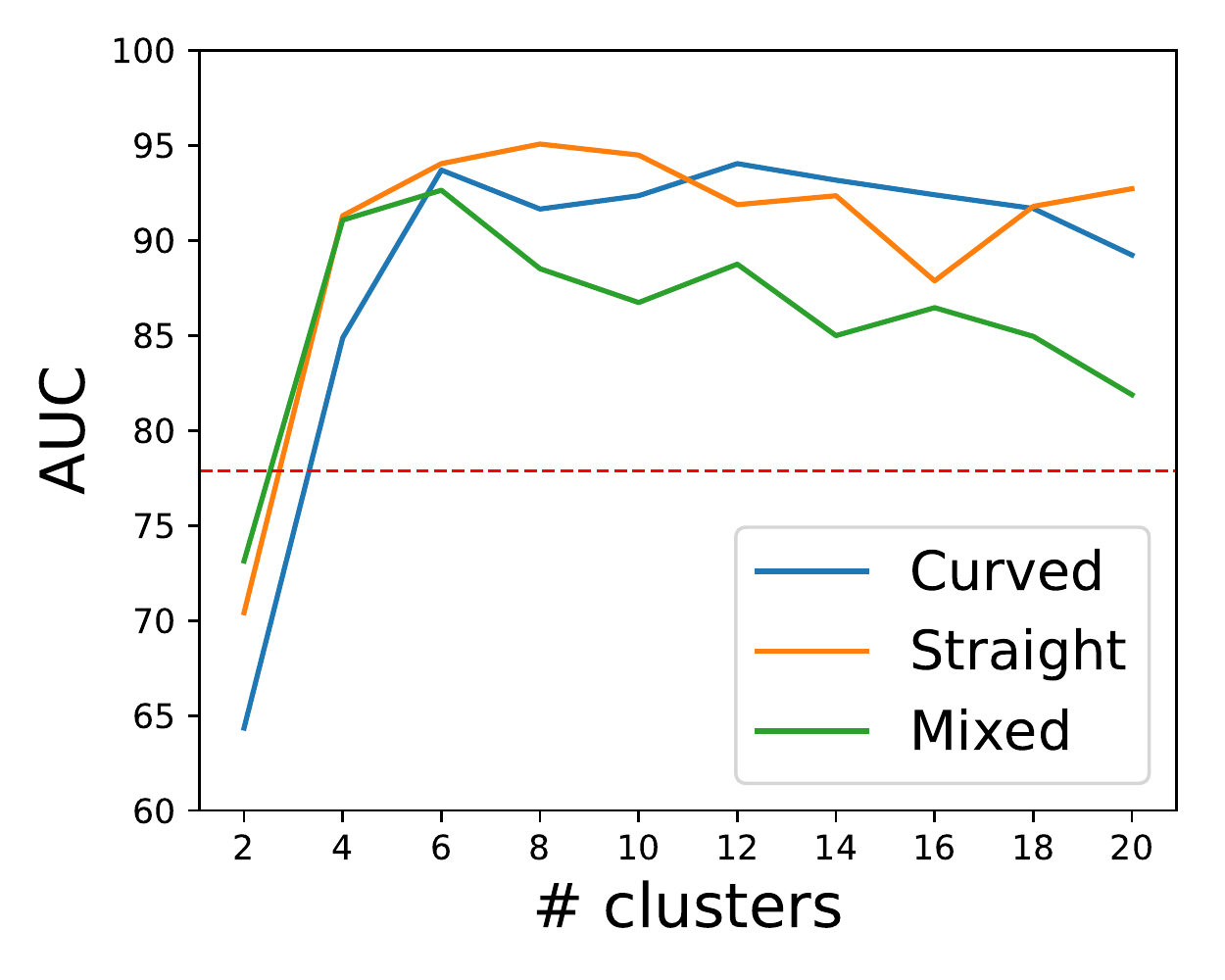}
    \caption{The effect of \# clusters of our scenarios.}
    \label{fig:ablation_cluster_num}
\end{subfigure}
\begin{subfigure}[b]{0.48\textwidth}
    \centering
    \includegraphics[width=\textwidth]{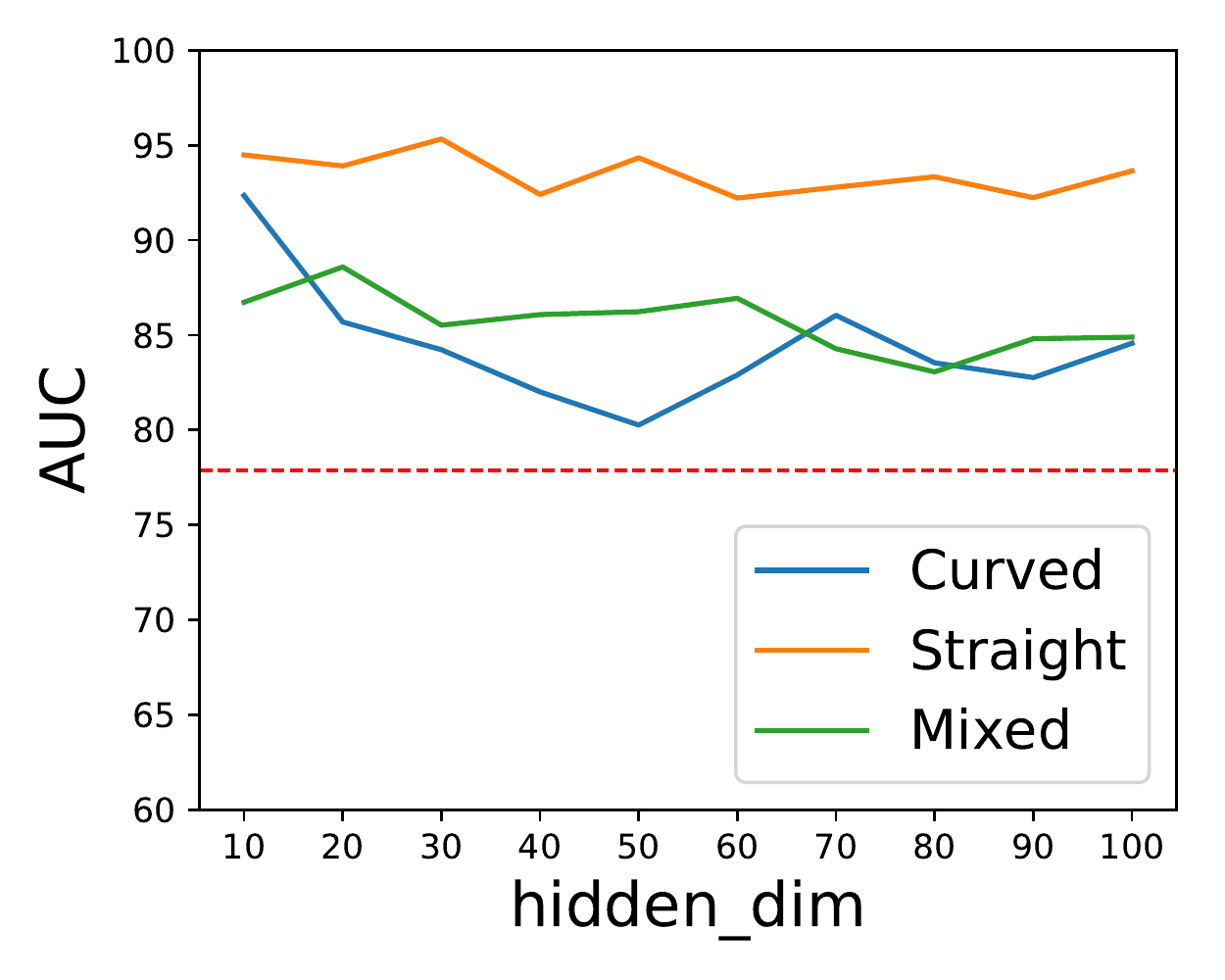}
    \caption{The effect of different hidden dimension sizes.}
    \label{fig:ablation_hidden_dim}
\end{subfigure}
\caption{Ablation study results. The red dash-line denotes the average scores of DeepSVDD on our three MNIST scenarios.
\textbf{(a)} With more than four clusters, our method achieves better performance compared to DeepSVDD. \textbf{(b)} We vary the hidden dimension sizes from 10 to 100. Our method shows consistent performance for all the hidden dimension sizes.}
\label{fig:ablation_study}
\end{figure}

\textbf{The effects of the hyper-parameters.} 
In our framework, we use feature extraction and clustering modules.
Since these modules are the basis of the self-labeling procedure, we analyze how the hyper-parameters of such modules affect the AD performance.
Figure~\ref{fig:ablation_study} shows the performance tendency when we change the number of the clusters or hidden dimension of feature $z$.

We vary the number of clusters from 2 to 20 and hidden dimension size from 10 to 100.
The red dashed line indicates the average scores of DeepSVDD in three scenarios.
If we set the number of cluster sizes to more than four, our method surpasses DeepSVDD by a huge margin.
This result implies that the robustness of CLAD to the rough self-labeling.
Our framework also shows the robustness with the change of the hidden dimension size $\mathbf{z}$; only marginal fluctuations are observed.

\section{Implementation Detail}\label{implementation_detail}

\textbf{Latent feature extraction.}
We use autoencoder-based architecture for this network.
Both encoder and decoder have five convolutional layers with increasing channel size followed by two linear layers.
We additionally apply dropout~\citep{srivastava2014dropout} to avoid overfitting.
The model training is done for 100 epochs using Adam~\citep{kingma2014adam} with a learning rate of 0.01.
Based on the model analysis, we set the number of clusters as 10 and the size of the hidden dimension as 100.

\textbf{Self-labeling via clustering.}
We adopt DEC~\citep{xie2016unsupervised} to self-assign labels to data samples.
In detail, we minimize KL divergence between the embedded data samples from encoder $E$ and the \textit{soft-alignments}.
When training this module, an SGD optimizer was used with a momentum of 0.9 and a learning rate of 0.01 for 100 epochs.

\textbf{Classifier for confidence-based AD.}
We use ResNet-18~\citep{he2016deep} as a classifier $F$.
We train this network using Adam~\citep{kingma2014adam} with a learning rate of 0.0001 for 100 epochs from scratch.
At the inference phase, we adopted the temperature scaling and input perturbation following ODIN~\citep{liang2017enhancing}.

\begin{table}[h]
\centering
\footnotesize
\begin{tabular}{c|c|c}
\hline
Dataset & Scenario & Class labels\\
\hline\hline
\multirow{3}{*}{MNIST} 
    & CUR (Curly)   & 0, 3, 8 \\
    \cline{2-3}
    & STR (Straight) & 1, 4, 7 \\
    \cline{2-3}
    & MIX (Mixed)          & 2, 5, 6, 9 \\
\hline\hline
\multirow{19}{*}{GTSRB}
    & \multirow{4}{*}{\begin{tabular}[c]{@{}c@{}}SPDL\\ (Speed Limit)\end{tabular}}
        & Speed limit (20km/h), Speed limit (30km/h), \\
        & & Speed limit (50km/h), Speed limit (60km/h), \\
        & & Speed limit (70km/h), Speed limit (80km/h), \\
        & & Speed limit (100km/h), Speed limit (120km/h) \\
    \cline{2-3}
    & \multirow{2}{*}{\begin{tabular}[c]{@{}c@{}}INST\\ (Driving Instruction)\end{tabular}}
        & No passing, No passing for vehicles over 3.5 metric tons, \\
        & & No vehicles, Vehicles over 3.5 metric tons prohibited \\
    \cline{2-3}
    & \multirow{6}{*}{\begin{tabular}[c]{@{}c@{}}WARN\\ (Warning)\end{tabular}}
        &  Right-of-way at the next intersection, General caution, \\
        & & Dangerous curve to the left, Dangerous curve to the right, \\
        & & Double curve, Bumpy road, Slippery road, \\ 
        & & Road narrows on the right, Road work, Traffic signals, \\
        & & Pedestrians, Children crossing, \\
        & & Bicycles crossing, Beware of ice/snow, Wild animals crossing \\
    \cline{2-3}
    & \multirow{3}{*}{\begin{tabular}[c]{@{}c@{}}DIRC\\ (Direction)\end{tabular}}
        & Turn right ahead, Turn left ahead, Ahead only, \\
        & & Go straight or right, Go straight or left, \\
        & & Keep right, Keep left, Roundabout mandatory \\
    \cline{2-3}
    & SPEC (Special Sign) & Priority Road, Yield, Stop, No entry \\
    \cline{2-3}
    & \multirow{3}{*}{\begin{tabular}[c]{@{}c@{}}REGN\\ (Regulation)\end{tabular}}
        & End of speed limit (80km/h), End of no passing, \\
        & & End of all speed and passing limits, \\
        & & End of no passing by vehicles over 3.5 metrics tons \\
\hline\hline
\multirow{2}{*}{CIFAR-10}
    & THG (Thing)  & Airplane, Automobile, Ship, Truck \\
    \cline{2-3}
    & LIV (Living) & Bird, Cat, Deer, Dog, Frog, Horse \\
\hline\hline
\multirow{10}{*}{\begin{tabular}[c]{@{}c@{}}Tiny-\\ ImageNet\end{tabular}} 
    & \multirow{2}{*}{\begin{tabular}[c]{@{}c@{}}ANML\\ (Animal)\end{tabular}} 
        & Golden retriever, Chihuahua, German shepherd, Labrador retriever, \\
        & & Standard poodle, Yorkshire terrier, Cougar/Puma, Persian cat \\
    \cline{2-3}
    & \multirow{2}{*}{\begin{tabular}[c]{@{}c@{}}ISCT\\ (Insect)\end{tabular}} 
        & Dragonfly, Roach, Bee, Grasshopper, \\
        & & Fly, Mantis, Monarch butterfly, Sulphur butterfly \\
    \cline{2-3}
    & \multirow{2}{*}{\begin{tabular}[c]{@{}c@{}}ISTM\\ (Instrument)\end{tabular}}
        & Water jug, Beer bottle, Tea pot, Pop bottle/Soda bottle, \\
        & & Beaker, Rugby ball, Volley ball, Pill bottle \\
    \cline{2-3}
    & \multirow{2}{*}{\begin{tabular}[c]{@{}c@{}}STRT\\ (Structure)\end{tabular}}
        & Triumphal arch, Suspension bridge, Fountain, Viaduct, \\
        & & Bannister, Steel arch bridge, Obelisk, Beacon \\
    \cline{2-3}
    & \multirow{2}{*}{\begin{tabular}[c]{@{}c@{}}VHCL\\ (Vehicle)\end{tabular}}
        & School bus, Trolly bus, Sports car, bullet train, \\
        & & Convertible, Tractor, Police van, beach wagon \\
\hline
\end{tabular}
\caption{Descriptions of sub-class scenario selection for each dataset.}
\label{tab:scenario_descriptions}
\end{table}

\section{Scenario Setting}
\label{sec:scenario_stup}

\noindent\textbf{MNIST.} We categorized the class labels into \{\texttt{Curly, Straight, Mixed}\} by the shape of the digit.
When training, we choose a single scenario as normal and the rest of the others as abnormal.
Note that the case where \texttt{Mixed} as normal is the most challenging since this category has similar features with other scenarios (\textit{e.g.} `6' versus `8').

\noindent\textbf{GTSRB.} This dataset is originally proposed for the traffic sign recognition task.
We choose the scenarios by following the subset as introduced in \cite{Stallkamp2012}.
With these subsets, we can simulate the abnormal cases from the driver or self-driving car perspective on various traffic signs.
Table~\ref{tab:scenario_descriptions} shows the overall scenarios and its containing class labels.

\noindent\textbf{CIFAR-10.} We divided the conditions into two simple scenarios as \{\texttt{Thing, Living}\} to mimic the real-world anomaly detection that can be used in general object recognition applications.

\begin{figure*}[t]
\centering
\setlength{\tabcolsep}{1em}
\begin{tabular}{ccc}
    \subfloat[CUR]{\includegraphics[width=0.12\linewidth]{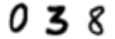}} &
    \subfloat[STR]{\includegraphics[width=0.12\linewidth]{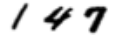}} &
    \subfloat[MIX]{\includegraphics[width=0.16\linewidth]{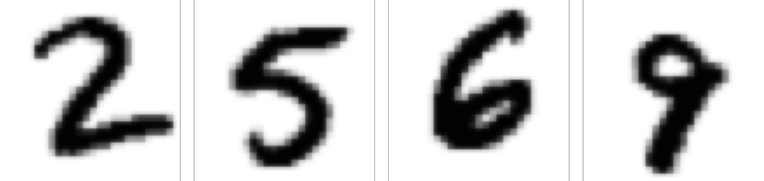}}
\end{tabular}
\begin{tabular}{cc}
    \subfloat[SPDL]{\includegraphics[width=0.4\linewidth]{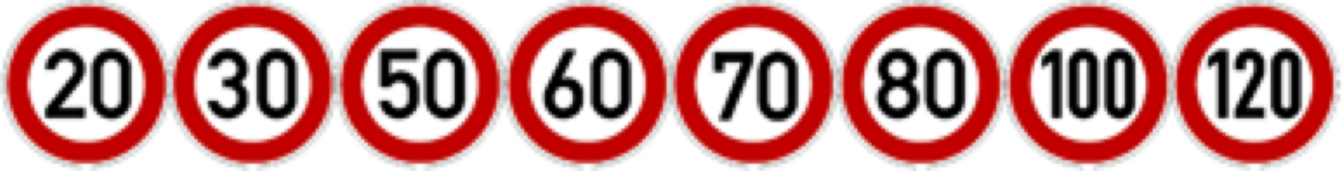}} &
    \subfloat[INST]{\includegraphics[width=0.2\linewidth]{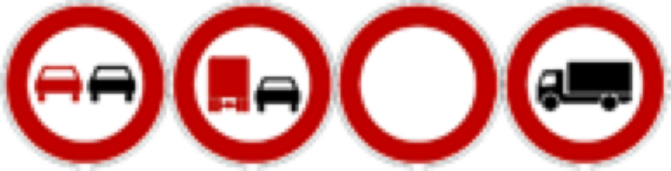}}
\end{tabular}
\begin{tabular}{c}
    \subfloat[WARN]{\includegraphics[width=0.75\linewidth]{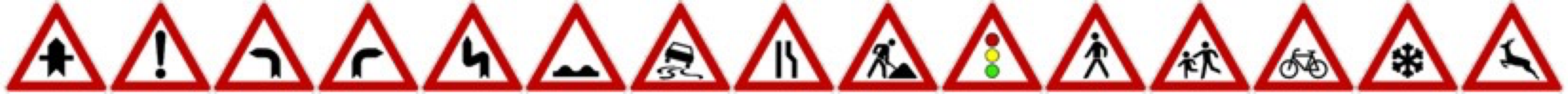}}
\end{tabular}
\begin{tabular}{ccc}
    \subfloat[DIRC]{\includegraphics[width=0.4\linewidth]{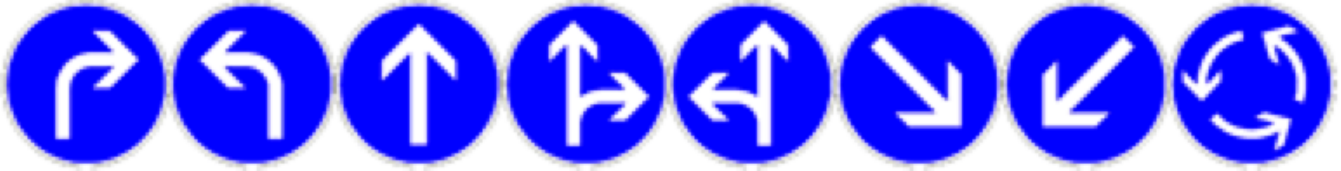}} &
    \subfloat[SPEC]{\includegraphics[width=0.2\linewidth]{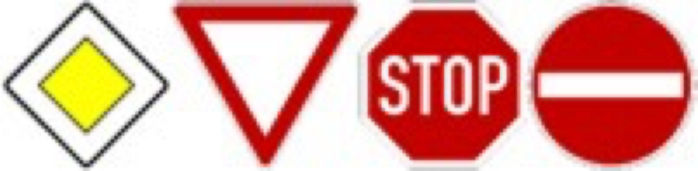}} &
    \subfloat[REGN]{\includegraphics[width=0.2\linewidth]{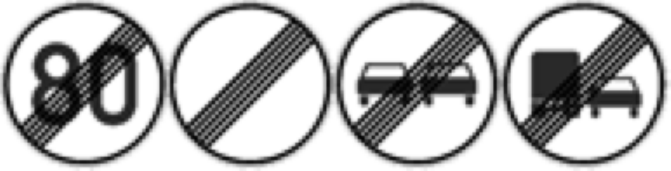}}
\end{tabular}
\begin{tabular}{cc}
    \subfloat[THG]{\includegraphics[width=0.24\linewidth]{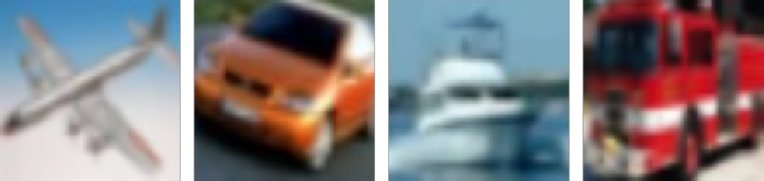}} &
    \subfloat[LIV]{\includegraphics[width=0.36\linewidth]{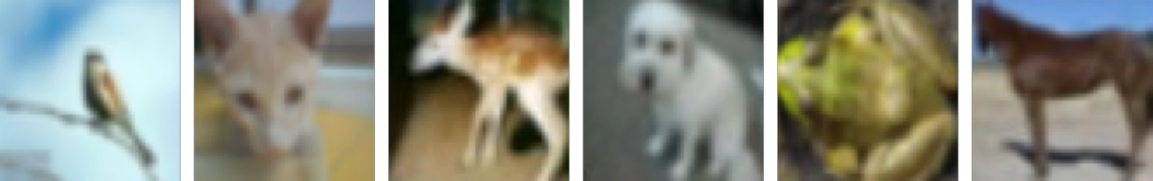}}
\end{tabular}
\begin{tabular}{ccc}
    \subfloat[ANML]{\includegraphics[width=0.24\linewidth]{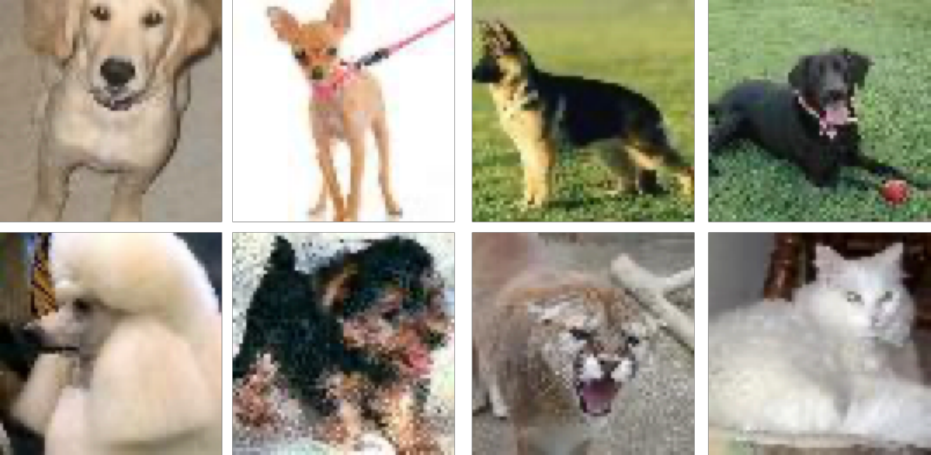}} &
    \subfloat[ISCT]{\includegraphics[width=0.24\linewidth]{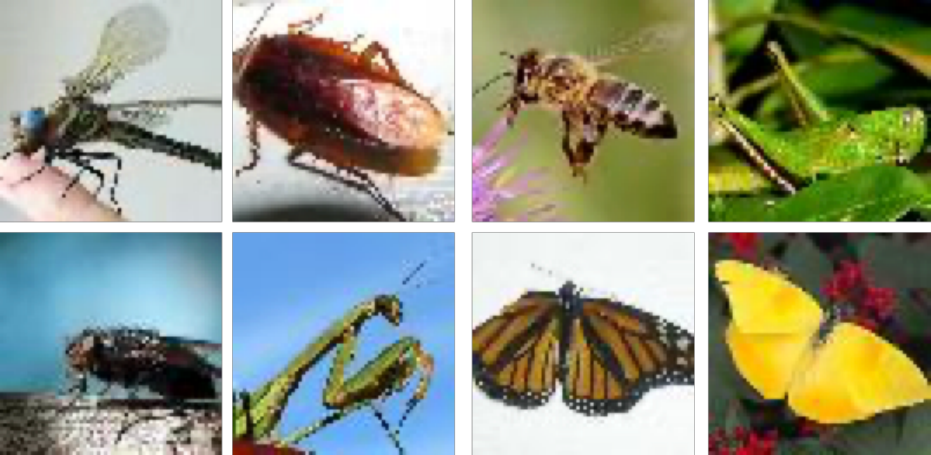}} &
    \subfloat[ISTM]{\includegraphics[width=0.24\linewidth]{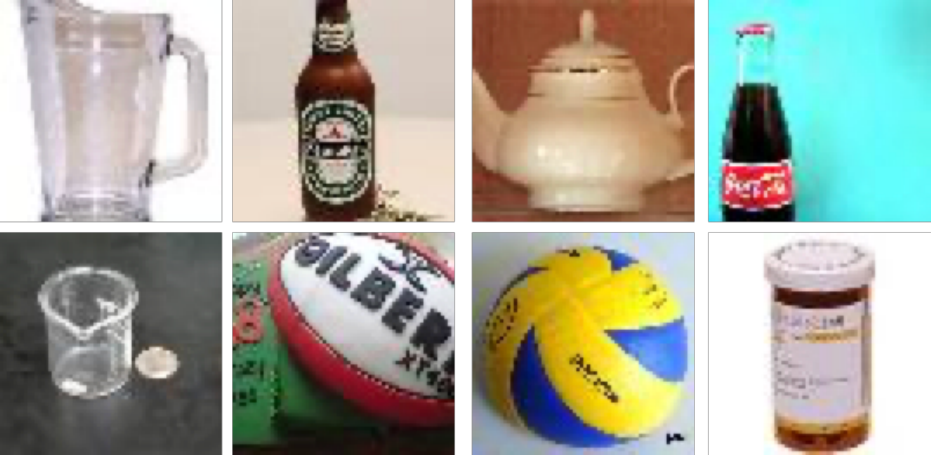}}
\end{tabular}
\begin{tabular}{cc}
    \subfloat[STRT]{\includegraphics[width=0.24\linewidth]{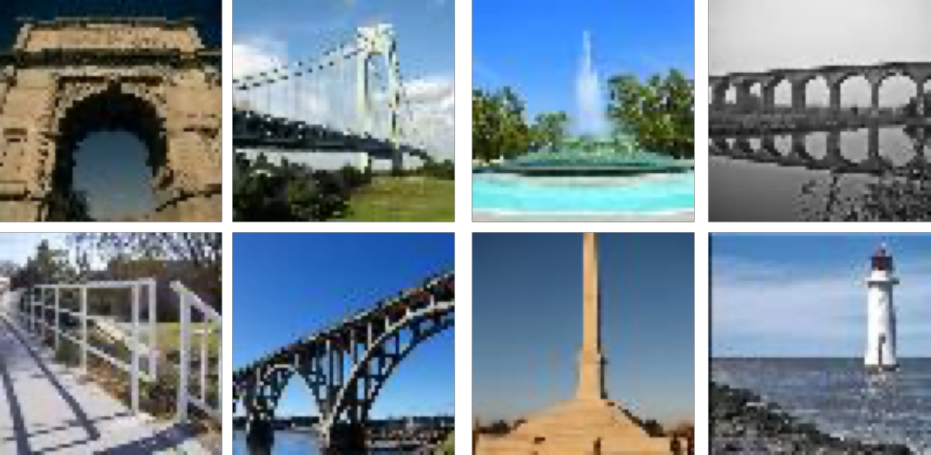}} &
    \subfloat[VHCL]{\includegraphics[width=0.24\linewidth]{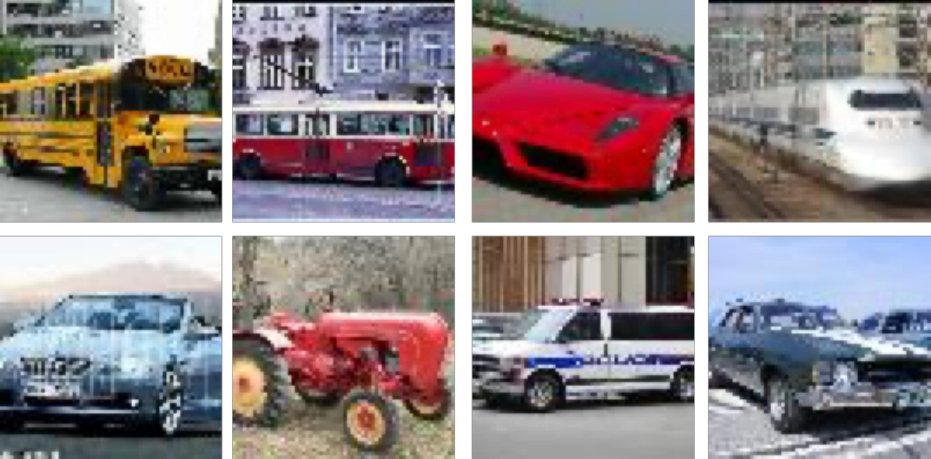}}
\end{tabular}
\caption{Representative images of the super-categories of each benchmark datasets: \textbf{(a-c)} MNIST.
\textbf{(d-i)} GTSRB.
\textbf{(j-k)} CIFAR-10.
\textbf{(l-p)} Tiny-ImageNet.}
\label{fig:representative_images}
\end{figure*}

\noindent\textbf{Tiny-ImageNet.} We first categorized class labels with representative subsets as \{\texttt{Animal, Insect, Instrument, Structure, Vehicle}\}.
Since the class labels of this dataset are annotated based on the WordNet~\citep{miller1995wordnet}, we selected the equal number of the classes for each scenario by referring to the same hierarchy as shown in Table~\ref{tab:scenario_descriptions} and the representative images for each scenario are shown in Figure~\ref{fig:representative_images}.

\end{document}